%% file: main.tex
\documentclass[10pt,twocolumn,letterpaper]{article}

\usepackage{iccv}              
\usepackage{graphicx}
\usepackage{booktabs}
\usepackage{multirow}
\usepackage{amsmath}
\usepackage{array}
\usepackage{adjustbox}

\usepackage{color, colortbl}
\definecolor{Gray}{gray}{0.93}
\newcolumntype{g}{>{\columncolor{Gray}}c}

\definecolor{Blue}{rgb}{0.9, 0.95, 1}
\newcolumntype{b}{>{\columncolor{Blue}}c}

\usepackage{amssymb}%
\usepackage{pifont}
\newcommand{\cmark}{\ding{51}}%
\newcommand{\xmark}{\ding{55}}%

\DeclareMathOperator*{\argmax}{arg\,max}
\newcommand{\modelname}{JEGAL\ }

\input{preamble}
\definecolor{iccvblue}{rgb}{0.21,0.49,0.74}
\usepackage[pagebackref,breaklinks,colorlinks,allcolors=iccvblue]{hyperref}

\title{Understanding Co-speech Gestures in-the-wild}

\author{Sindhu B Hegde\thanks{equal contribution}, \quad
K R Prajwal\footnotemark[1], \quad
Taein Kwon, \quad
Andrew Zisserman\\
Visual Geometry Group, Dept.\ of Engineering Science, University of Oxford\\
{\tt\small \{sindhu, prajwal, taein, az\}@robots.ox.ac.uk}\\
\small{\url{https://www.robots.ox.ac.uk/~vgg/research/jegal}}
}

\begin{document}
\maketitle

\begin{abstract}
Co-speech gestures play a vital role in non-verbal communication. In this paper, we introduce a new framework for co-speech gesture understanding in the wild. Specifically, we propose three new tasks and benchmarks to evaluate a model's capability to comprehend gesture-speech-text associations: (i) gesture based retrieval, (ii) gesture word spotting, and (iii) active speaker detection using gestures. We present a new approach that learns a tri-modal video-gesture-speech-text representation to solve these tasks. By leveraging a combination of global phrase contrastive loss and local gesture-word coupling loss, we demonstrate that a strong gesture representation can be learned in a weakly supervised manner from videos in the wild. Our learned representations outperform previous methods, including large vision-language models (VLMs). Further analysis reveals that speech and text modalities capture distinct gesture related signals, underscoring the advantages of learning a shared tri-modal embedding space. The dataset, model, and code are available at: \url{https://www.robots.ox.ac.uk/~vgg/research/jegal}.
\end{abstract}

\section{Introduction}
\label{sec:intro}

\begin{figure}[ht]
\includegraphics[width=\linewidth]{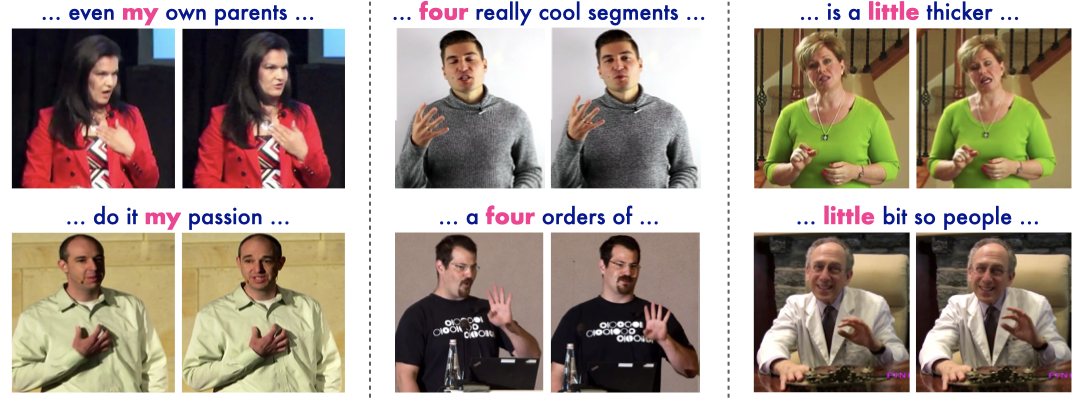}
    \caption{Co-speech gestures supplement the spoken language -- we show examples for six phrases here, with common words. Learning to associate gestures with the uttered phrases is essential for a holistic understanding of human communication.}
  \label{fig:teaser}
  \vspace{-10pt}
\end{figure}

Humans gesture when they talk -- gesturing is an integral part of human communication, together with speech and facial expressions. Gestures can vary from {\em beats} -- two phase hand movements (up/down, left/right etc) that emphasize particular words or phrases and match the rhythm of the speech, but do not carry semantic content -- to {\em
iconic} and {\em deictic} gestures that are representational and illustrate the {\em content} of the speech~\cite{andric2012gesture,mcneill1992hand}. For example, hands and arms moving apart can accompany a speech segment indicating that something is ``huge'', or as illustrated in Fig~\ref{fig:teaser}, an inward pointing gesture to depict the uttered word ``my''. 

Non-verbal communication accounts for $55\%$ of overall communication\footnote{\footnotesize{https://online.utpb.edu/about-us/articles/communication/how-much-of-communication-is-nonverbal/}}, highlighting the need for machines to  understand non-verbal gestural elements in order to have a holistic understanding of human communication. A clear application is enriching human-computer interaction (HCI) through gestures, and this requires machines to comprehend the semantics of the user's hand gestures. Another application is to detect if a person is speaking based on their gestures, or spot specific words or phrases in a video based on gestures alone. More generally, being able to recognize gestures and determine their semantic and temporal alignment with speech enables human communication to be studied at scale~\cite{kendon2004gesture}.

In this paper, our objective is to learn and evaluate co-speech gesture representations. To this end, we propose
three tasks and evaluation benchmarks that act as a proxy for assessing real world applications: (1) gesture based cross-modal retrieval, (2) gesture word spotting, and (3) active speaker detection via gestures. We perform large-scale training featuring $\approx 7000$ speakers and evaluate on in-the-wild videos from the AVSpeech dataset~\cite{Erphat_LookingToListen_2018}.

All three tasks require models to learn to associate gesture clips with speech segments or their corresponding textual transcripts. For this, we propose a model termed \textbf{J}oint \textbf{E}mbedding space for \textbf{G}estures, \textbf{A}udio, and \textbf{L}anguage ({\bf JEGAL})\footnote{\footnotesize{JEGAL, known as Zhuge Liang in Chinese, was a prominent historical figure from China's Three Kingdoms period and is regarded as a symbol of wisdom.}} that facilitates matching gestures to words and phrases in the accompanying speech. The matches can be based on the style of the speech (intonation, stress, prosody) or the semantic content of the phrase or particular words. However, learning a rich joint gesture-audio-language embedding space is a very challenging task. The associations between gestures and speech are typically sparse and ambiguous, with a high degree of variability across speakers. Usually, only a few of the spoken words are clearly gestured. Additionally, the same sentence can be gestured very differently in different contexts and by different people. Gestures also depend on the speaker's emotion, culture, and social scenario (formality, private vs.\ public, with friends or strangers, etc.). Furthermore, some types of gestures, such as beat gestures, carry no semantic information, resulting in no direct mapping from the gesture to words. The sparse and weak cross-modal correlations makes gesture representation learning a very unique research problem. 

We make three key design choices in our approach that result in a strong tri-modal gesture representation. To start with, we learn gesture video representations from large-scale {\em weak} cross-modal supervision. The supervision is weak because we only use phrase-level speech audio and transcripts -- since we do not have any information on which words in the speech are gestured for videos collected `in-the-wild'. Second, we obtain cross-modal supervision in the form of both audio and the corresponding text transcript (in Section~\ref{subsec:speechVStext}, we demonstrate that speech and text modalities capture complementary gesture-related signals). Third, we introduce a new gesture-word alignment and spotting loss that explicitly encourages learning of word-level correspondences.

To summarize, we make the following contributions: (i) we propose a new framework for co-speech gesture understanding with three new tasks and evaluation benchmarks; (ii) we learn a joint tri-modal embedding space in a weakly-supervised manner with a combination of global phrase-level objective, and a local word-level gesture coupling loss; (iii) we demonstrate that the learned \modelname representation performs on par with vision-language foundation models on three gesture-centric tasks and is useful for practical applications.

\section{Related Work}

\begin{figure*}[ht]
\includegraphics[width=\textwidth]{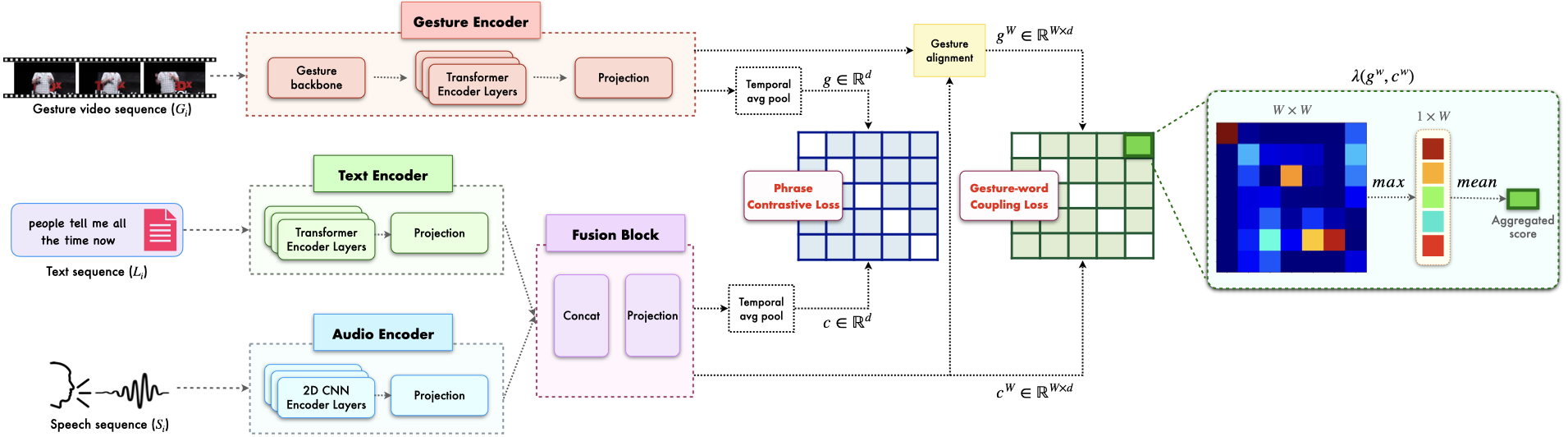}
  \vspace{-14pt}
  \caption{
  The \textbf{\modelname} architecture. The three input modalities (video, text, speech) are each encoded with a modality-specific encoder, followed by a fusion block to merge speech and text representations. The encoder outputs are average-pooled to obtain global (phrase-level) gesture and speech-text embeddings. During training, these provide the inputs for the global `phrase contrastive loss'. The gesture alignment module aggregates the relevant video frames to obtain a local word-level gesture embedding for each speech-text word. During training, these provide the inputs for the local `gesture-word coupling loss'. The two losses encourage the learning of global and local correspondences between the three modalities.}
  \label{fig:pipeline}
  \vspace{-14pt}
\end{figure*}

\noindent \textbf{Spectrum of Human Gestures.}  Gestures can be classified broadly into four major classes~\cite{mcneill1992hand}. Emblematics convey a clear symbolic meaning, e.g.\ a ``thumbs-up''. Iconic gestures are used to convey meaning co-occurring with the articulated speech, e.g.\ ``revolving door'' is accompanied with the hands moving in a circular motion. Deictic gestures are pointing gestures using the index finger. The most common are the ``beat'' gestures which are co-speech gestures that are temporally aligned with the prosodic characters of human speech. They are prominent on lexically stressed syllables. Past works~\cite{wagner2014gesture,kendon2004gesture} have studied how these gesture classes relate to the speech.  
Several works have attempted to recognize hand, head, and facial emblematic gestures~\cite{freeman1995orientation,darrell1996task}. Recognizing deictic gestures can help in identifying the referring object being pointed to in a conversation, an essential part of human-robot interaction~\cite{matuszek2014learning}. The other two gesture classes, i.e., \ beat and iconic, are quite diverse and are difficult to associate with a well-defined set of words or hand movements. This work takes a data-driven approach to learn gesture representations with the help of speech and natural language.
\vspace{2pt}

\noindent \textbf{Co-speech Gesture Understanding.} Spoken discourse consists of multiple streams of information: language, lip movements, facial expressions, and hand gestures. As opposed to the advancements in the other streams~\cite{radford2023robust,Prajwal22,barrault2023seamless}, co-speech gesture understanding is a relatively under-explored area. One possible reason for this could be that gestures are only sparsely correlated with the speech. As a result, some of the works have resorted to developing models for specific cases where the gestures are clear, e.g.\ weather narration~\cite{sharma2000exploiting,kettebekov2003improving}. Some works~\cite{molchanov2016online,morency2007latent,ghaleb2024co} have tried to detect and recognize gestures in laboratory settings while using multiple stereo and IR cameras. Recently, GestSync~\cite{Hegde23} learns gesture representations by solving for cross-modal synchronization with speech. This objective can lead to the model capturing low-level associations rather than high-level semantics, leading to poor performance on tasks like retrieval, and word spotting.
Our work is the first one to learn gesture representations which capture the semantics, style and also learn word-level associations.
\vspace{2pt}

\noindent \textbf{Understanding Gestures in Sign Language.} Sign language understanding and recognition is another body of work where models need to understand and associate gestures to words and phrases to solve tasks like sign recognition~\cite{Albanie2021bobsl,Prajwal22a,raude2024,zuo2022c2slr,min2021visual,zhou2020spatial,camgoz2020sign}, sign language retrieval~\cite{cheng2023cico,duarte2022sign}, sign language translation~\cite{tarres2023sign,yin2023gloss,camgoz2018neural,camgoz2020sign} and sign language production~\cite{stoll2020text2sign,saunders2020progressive,saunders2022signing}. Sign language gesture understanding is quite different compared to co-speech gesture understanding. In sign language, text transcription is a~\textit{translation} of what is being signed/gestured. Thus, the words in the text can be a summary or paraphrasing of what is being gestured, with even a mismatch in the temporal ordering. In co-speech gestures, the speaker is the gesturer, and the gestures are being made by the speaker to directly accompany each word he (she) utters. Hence, these two tasks require very different approaches.
\vspace{2pt}

\noindent \textbf{Co-speech Gesture Generation.} Several works have focused on generating natural gestures that match a given speech segment. This task has the advantage of being ``freely supervised'' -- large-scale datasets can be curated for this task with almost no manual effort as it only requires unlabeled videos of people talking. Speech2Gesture~\cite{speech2gesture} trains speaker-specific models to generate hand skeleton motion for a given speech segment. Recent papers~\cite{trimodal,livelyspeaker,ao2023gesturediffuclip} have moved towards more speaker-independent approaches while also using text to obtain strong semantic supervision. In particular, GestureDiffuCLIP~\cite{ao2023gesturediffuclip} learns a joint gesture-text embedding to improve gesture generation. As will be seen in the results, one clear distinction from the work presented in this paper is that~\cite{ao2023gesturediffuclip}  does not learn word-level correspondences, which makes a significant difference to the gesture understanding tasks that we evaluate.

\noindent\textbf{Gesture Recognition Datasets.}
ChaLearn ConGD and IsoGD~\cite{Wan16chalearn} are two gesture recognition datasets, providing
benchmarks for the ChaLearn challenges. 
However, these datasets are of people using gestures for a task, (e.g. playing a game or controlling an appliance), and
are not suitable for learning or evaluating co-speech gestures.
Montalbano II~\cite{escalera2013multi} is another dataset covering gestures from a vocabulary of $20$ Italian sign gesture categories. Again, this is not suitable for our task.

\vspace{5pt}
\noindent \textbf{Learning Video Representations.} 
Representation learning in videos~\cite{tong2022videomae, wang2023videomae,oquab2023dinov2, qian2021spatiotemporal} has gained significant attention driven by the availability of large-scale video datasets. 
Learning video representations from text offers a promising advantage by incorporating interpretability via language. Recent works on vision language representation learning~\cite{clip, flamingo, li2023blip, videobert, wang2022internvideo} highlight this potential. On the other hand, since audio is naturally paired with video, other studies~\cite{lee2021acav100m,shi2022learning} have explored learning video representations from audio. 
More recently, multimodal approaches have emerged for video representation learning. LanguageBind~\cite{languagebind} utilizes depth, infrared, audio, and video to enhance video representations, while Video-LLaMA~\cite{zhang2023video} learns video representations from free-form text and audio.
Following these successes, our work aims to leverage multimodal data - video, audio and text - to advance the understanding of gestures.

\section{Method}
\vspace{-5pt}
Our goal is to learn co-speech gesture representations from speech and text supervision. Given a dataset {\it G, S, L} of gesture clips, the accompanying speech segments and their corresponding transcriptions, our goal is to learn gesture representations that capture the rich semantics (from text) and utterance style (from speech) of what is being spoken. 

\subsection{Overview}
\modelname learns gesture representations by solving two multimodal contrastive objectives between the gesture video and the two other modalities, i.e., speech and text. Each of the three modalities are first encoded using separate encoders $\mathbb{G}, \mathbb{S}, \mathbb{L}$ to get modality-specific embeddings $\mathbf{g^T} \in \mathbb{R}^{T\times d}, ~\mathbf{s^w} \in \mathbb{R}^{W\times d/2}, ~\mathbf{l^w} \in \mathbb{R}^{W\times d/2}$ to obtain frame-level ($T$) and word-level ($W$) representations. The speech and text embeddings are fused into joint speech-text embeddings $\mathbf{c^w} \in \mathbb{R}^{W\times d}$ as depicted in Fig~\ref{fig:pipeline}. 

We learn these representations using (1) a {\em global phrase-level contrastive objective},  and (2) a {\em local gesture-word coupling loss}. The first one encourages the model to learn global semantics to match a gesture clip to a speech/text segment. The second objective enforces the model to find the strongest word-level matches between the gesture clip and the other two modalities. We describe our architecture and loss functions in detail below. 

\subsection{Gesture Encoder}
\noindent \textbf{Gesture backbone.} Given a gesture clip $G \in (T, h, w, 3)$ of $T$ frames, we encode it using a stack of 3D convolution layers, similar to previous audio-visual networks~\cite{Afouras20b,Prajwal22,Prajwal22a}, where the first layer has a temporal receptive field of $5$ frames to capture the motion information. We obtain a sequence of $T$ visual feature vectors of dimension $d$. These feature vectors are further encoded using a stack of Transformer encoder layers to get $\mathbf{f_g} \in \mathbb{R}^{T\times d}$. We initialize the backbone weights from GestSync~\cite{Hegde23} and keep it frozen.

\noindent \textbf{Gesture head.} We use a Transformer encoder followed by a projection layer to get the gesture embeddings, $\mathbf{g^T} \in \mathbb{R}^{T\times d}$. 

\subsection{Text Encoder}
\noindent \textbf{Text backbone.} Given a text transcription $L$ corresponding to the gesture video clip, we use the final layer outputs from a pre-trained bi-directional language model, multilingual Roberta XLM-Base~\cite{roberta-xlm} to obtain text representations. The output of the text backbone is a sequence of sub-word feature vectors, $f_l$.

\noindent \textbf{Text head.} The text head is similar to the gesture head that uses a stack of Transformer layers. The sub-word embeddings $f_l$ are encoded and projected to get the final sub-word embeddings $l^{sw}$ of feature dimension $d/2$ each. We aggregate these sub-word tokens to word-level tokens later in the fusion block.

\subsection{Audio Encoder}
Given a speech waveform $S$, we convert it into melspectrograms which is encoded using a stack of 2D-CNN layers following previous works~\cite{Chung16a,Afouras20b}. The output of the audio encoder is a sequence of $T'$ speech feature vectors, $\mathbf{s} \in \mathbb{R}^{T'\times d/2}$.

\subsection{Fusion Block}
Before we fuse the speech and text embeddings, we aggregate the text and audio embeddings to obtain word-level feature vectors. We average the sub-word embeddings for each word to get word-level text embeddings $\mathbf{l^w} \in \mathbb{R}^{W\times d/2}$. 
Using the start and end times of each word, we average the speech features for each word to get word-level features $\mathbf{s^w} \in \mathbb{R}^{W\times d/2}$.
We fuse the speech and text features by concatenating along the feature dimension to get joint word-level representations, $\mathbf{c^w} \in \mathbb{R}^{W\times d}$.

\subsection{Gesture-Word Alignment}
\label{subsec:alignment}
The word boundaries are aligned with the speech, but not necessarily with the gestures in the video. The gestures can be longer or shorter than the window in which the word is uttered, and can also be offset. To handle this discrepancy, we propose an attention-based pooling mechanism to obtain the gesture embedding corresponding to the word. We first pad the speech-based word start-end times with $p=10$ video frames on either side. Let $S, E$ be the start and end frames for the padded window for the word $c^{w_i}$. We obtain the word-level gesture embedding, $g^{w_i}$ by using the word embedding $c^{w_i}$ for attention-pooling over the extended temporal interval of the word:
\begin{equation}
    g^{w_i} = \sum_{j=S}^{E} \left( \frac{\exp(\gamma \cdot g^{T_j} \cdot c^{w_i})}{\sum_{j=S}^{E} \exp(\gamma \cdot g^{T_j} \cdot c^{w_i})} \right) \cdot g^{T_j}
\end{equation}

\subsection{Training Objective}
We only have reliable supervision at the phrase level — meaning we know that a specific text or speech segment corresponds to a gesture clip. However, we do not know which individual words are gestured. Keeping this in mind, we employ two loss functions.

\vspace{2pt}
\noindent \textbf{Global Phrase Contrastive Loss.} To obtain the global phrase embeddings, we average pool the speech-text embeddings and the video frame embeddings to give us $\mathbf{c}$ and $\mathbf{g}$ respectively. Given a batch of $N$ samples, we employ the contrastive Info-NCE loss~\cite{oord2018representation} to encourage similarity between the $N$ positive triplets, and dissimilarity between the $N^2 - N$ negative triplets:

\vspace{-15pt}
\begin{equation}
    \mathcal{L}_{seq} = -\frac{1}{N} \sum_{i=1}^{N} \left( \log \frac{\exp(\gamma \cdot cos(g_i, c_i))}{\sum_{j=1}^{n} \exp(\gamma \cdot cos(g_i, c_{j}))} \right)
\end{equation}

\noindent where $\gamma$ is the temperature and $cos$ is the cosine similarity.

\noindent \textbf{Local Gesture-word Coupling Loss.} 
Pooling the word-level representations to compute the global phrase loss can lead to weak gesture-word associations (Table~\ref{tab:loss}, row $1$). However, directly training to match gestures to words is not possible -- very few words are gestured in a given phrase, and we do not know which of them are. Thus, we devise a new strategy to learn word-level correspondences using phrase-level supervision. Given a pair of word-level gesture and speech-text ($\mathbf{g^w}, \mathbf{c^w}$) embeddings, we first find the closest gesture $g^{w_j}$ for each speech-text word $c^{w_i}$. Our hypothesis is that a matching ($\mathbf{g^w}, \mathbf{c^w}$) will have a higher number of strong word couplings than a non-matching pair. With this idea, we define the following scoring function and the gesture-word coupling loss:
\vspace{-5pt}
\begin{equation}
    \lambda(g^w_n, c^w_n) = \frac{1}{W} \sum_{i=1}^{W} \max\limits_{j=1,2..W} cos(g^{w_i}_n, c^{w_j}_n)
\end{equation}
\vspace{-2pt}
\begin{equation}
    \mathcal{L}_{couple} = -\frac{1}{N} \sum_{i=1}^{N} \left( \log \frac{\exp(\gamma \cdot \lambda(g^w_i, c^w_i))}{\sum_{j=1}^{N} \exp(\gamma \cdot \lambda(g^w_i, c^w_{j}))} \right)
\end{equation}

The gesture-word coupling loss simply maximizes $\lambda$ for matching gesture-speech-text samples while minimizing $\lambda$ for the negative ones. In other words, the model is encouraged to find more strong word-level couplings for positive gesture-speech-text phrases in the batch.

\noindent
Our final loss function is a weighted sum of the two losses:

\vspace{-5pt}
\begin{equation}
    \mathbb{L} = \beta\ \cdot\ \mathcal{L}_{seq}\ + (1 - \beta)\ \cdot\ \mathcal{L}_{couple}
\end{equation}

\subsection{Implementation Details}
We now describe the essential implementation details, more details can be found in the supplementary material.

\noindent \textbf{Training data.} We train our model, JEGAL, on triplets of gesture clips, speech segments, and text transcriptions. For the gesture frame inputs, we resize the frames to $270\times480$ pixels. We extract melspectrograms with a hop length of $10$ms. The word-aligned text transcriptions  are tokenized into wordpiece tokens. Using the start-end time of the word boundaries, we randomly sample a video clip between $2-10$ seconds in length. 

\noindent \textbf{Modality drop.} In order to encourage the model to learn both speech and text representations equally well, we randomly set one of these modality inputs to zero $50\%$ of the time. This is commonly done in audio-visual speech recognition models~\cite{shi2022learning,Afouras18b}. This also allows us to use only one modality (speech or text) during inference, if necessary.

\noindent \textbf{Model hyper-parameters.} For the text and gesture heads, we set the number of Transformer layers to $3$ and $6$ respectively. The Transformer uses a hidden dimension of $512$ and a feed-forward dimension of $2048$ with $8$ attention heads. 

\noindent \textbf{Training hyper-parameters.} We use the AdamW optimizer~\cite{adamw} with a learning rate of $5e^{-5}$, weight decay of $1e^{-4}$ and betas $(0.9, 0.98)$. We reduce the learning rate by a factor of 5 when the validation performance does not improve for 2 epochs.

\subsection{Training Datasets}
We train our model using the following datasets: (i) PATS~\cite{speech2gesture}, and (ii) a subset of the MultiVSR dataset~\cite{multivsr}. 
The dataset specifics are outlined in Table~\ref{tab:dataset}. 
PATS~\cite{speech2gesture} is a publicly available video dataset from $25$ speakers sourced from diverse platforms such as lectures, talk-shows, YouTube, and televangelists. 
The subset from the MultiVSR dataset is composed of $556$ hours of interviews, narrations, and talks spanning a broad spectrum of speakers and a rich vocabulary.

\noindent
\textbf{Pre-processing:} 
We resample all videos to $25$ FPS, and the speech to 16kHz. We leverage WhisperX~\cite{Bain23} in cases where datasets lack word-aligned text transcripts. Additionally, using the $L2$ distance between consecutive frame body keypoints, we filter out samples with minimal gesture activity. We also make sure to mask out the face region to avoid leakage from lip movements. Table~\ref{tab:dataset} presents the final statistics of all the datasets.

\begin{table}[tb]
  \caption{We train and evaluate on multiple datasets consisting of $720$ hours of gesture clips comprising $7000+$ speakers. For evaluation, we curate task-specific benchmarks from the publicly available AVSpeech~\cite{Erphat_LookingToListen_2018} dataset.}
  \label{tab:dataset}
  \centering
  \small
  \setlength{\tabcolsep}{3pt}
  \vspace{-10pt}
  \begin{tabular}{l|c|c|c|c|c}
    \toprule
    \multirow{2}{*}{\textbf{Dataset}} & \multirow{2}{*}{\textbf{split}} & \multirow{2}{*}{$\#$ \textbf{hours}} & \multirow{2}{*}{$\#$ \textbf{spk.}} & \textbf{avg. clip} & \multirow{2}{*}{$\#$ \textbf{videos}} \\
    & & & & \textbf{duration (s)} &\\
    \midrule
    
    \multirow{1}{*}{PATS~\cite{speech2gesture}} & train & 162.3 & 24 & 11.37 & 51390\\
    MultiVSR~\cite{multivsr} & train & 556.1 & 6934 & 15.31 & 130510\\

    \hline
    \hline
    \textbf{Combined} & train & 718.4 & 6958 & 14.2 & 181900\\
    \hline
    \hline
    \rowcolor{Blue}
    \textbf{AVS-Ret} & test & 0.31 & 404 & 2.27 & 500 \\
    \rowcolor{Blue}
    \textbf{AVS-Spot} & test & 0.38 & 384 & 2.76 & 500\\
    \rowcolor{Blue}
    \textbf{AVS-Asd} & test & 0.44 & 398 & 3.15 & 500\\
    
  \bottomrule
  \end{tabular}
  \vspace{-10pt}
\end{table}

\section{Downstream Tasks and Evaluation}
We describe our newly curated evaluation benchmarks and the different downstream tasks to evaluate the quality of our learned gesture representations. The first is cross-modal retrieval, the second is spotting gestured words, and the third is active speaker detection. Note that in all the tasks, while we use the joint speech-text embedding, we can obtain uni-modal scores by inputting zeros to omit a modality during inference.

\subsection{Cross-modal Retrieval} 
Given a gallery of gesture-speech-text samples, the task is to retrieve a gesture clip given a speech segment and/or text and vice-versa. Concretely, given a speech or text as query, we obtain a speech-text embedding, $c \in \mathbb{R}^d$ and rank the gesture embeddings $g \in \mathbb{R}^d$ in the gallery by cosine similarity, highest being at the top. We do the same process for the gesture to speech-text retrieval as well.

Retrieving relevant gestures for a text or speech segment enables several practical applications. For digital avatars, we can retrieve most plausible hand gesture clips to accompany what the avatar is speaking, leading to a more immersive and engaging experience. In gaming applications, given a database of gesture sequences, the developer can automatically select the most relevant gestures to go with the in-game dialogues. Gestures can assist in language learning~\cite{macedonia2011impact} by improving word-level memory retention (e.g.\ eat, kick, clap). Language teaching apps will be able to retrieve gesture clips for sentences to improve the speed of foreign language learning.

\subsection{Gesture Word Spotting}
Given a gesture clip with the accompanying speech/text segment and a word of choice from this segment, the goal is to localize the word in the gesture clip. Concretely, we obtain word-level speech-text ($c^w$) embeddings and frame-level gesture embeddings, $g^T$. To localize the i-th word, $c^{w_i}$, we compute the cosine similarity of the word embedding with all the gesture frame embeddings. The localization of the word in the video is simply obtained by keeping only the locations with similarity scores $\ge \delta=0.5$.

Spotting can be useful to enhance transcriptions by supplementing the plain words with stress and emotion labels. Another application would be to create word-level gesture databases, e.g.\ a thousand different ways the word ``big" is gestured by people all over the world, which will be useful for language and communication analysis.

\subsection{Active Speaker Detection}
Given gesture clips of $P$ different speakers, and a speech ($S$) and/or text segment ($T$), the goal is to predict the active speaker $A$ who is uttering the queried speech/text. To do this, we extract the sequence-aggregated gesture features $g_{i} \in \mathbf{R}^{d}, i \in {1, 2, ..., P}$ for each of the $P$ clips. Given the query speech or text, we obtain the speech-text feature, $c$. The active speaker $A$ is the one whose gesture and speech-text cosine similarity is maximum:

\vspace{-10pt}
\begin{equation}
  A = \argmax_{i \in {1, 2, ..., S}}\ cos(c,\ g_i)
\end{equation}
\vspace{-8pt}

The majority of audio-visual models, encompassing tasks like speech recognition, generation, and translation, primarily operate on inputs containing a single speaker. Thus, there arises a necessity to identify the speaker within a video segment. To determine the active speaker in a multi-speaker scenario, previous works~\cite{Afouras18,Rahimi22} have shown the benefits of resorting to the face for lip-sync with the audio, and text subtitles when the audio is corrupted. We extend this thread even further. What happens if the lip region is occluded or unclear? Another important use-case is privacy preserving~\cite{xu2021privacy} active speaker detection: what if the active speaker detection needs to be done without leaking the face identity of the speaker? We show that we can successfully do this -- with very little identity information, i.e.\ by only using the hand gestures, we can determine who is speaking.

\begin{table*}[h]

  \caption{Cross-modal retrieval performance on the AVS-Ret benchmark (Sec~\ref{sec:test_datasets}).~\modelname outperforms the baselines by a large margin.}
  \label{tab:t2g}
  \centering
  \setlength{\tabcolsep}{4.8pt}
  \vspace{-5pt}
  \begin{adjustbox}{width=1.0\textwidth,center}
  \begin{tabular}{l|c|c|c|c|c|c|c||c|c|c|c|c}
    \toprule

    \multirow{2}{*}{\textbf{Method}} & \multicolumn{2}{c|}{\textbf{Mod.}} & \multicolumn{5}{c||}{\textbf{Speech-text to Gesture retrieval}} & \multicolumn{5}{c}{\textbf{Gesture to Speech-text retrieval}}\\\cline{2-13}
    & \textbf{T} & \textbf{A} & \textbf{R@5} $\uparrow$ & \textbf{R@10}$\uparrow$ & \textbf{R@25}$\uparrow$ & \textbf{R@50}$\uparrow$ & \textbf{MR} $\downarrow$ & \textbf{R@5} $\uparrow$ & \textbf{R@10}$\uparrow$ & \textbf{R@25}$\uparrow$ & \textbf{R@50}$\uparrow$ & \textbf{MR} $\downarrow$ \\
    
    \midrule
    Random & \cmark & \cmark & 1.00 & 2.00 & 5.00 & 10.00 & 250 & 1.00 & 2.00 & 5.00 & 10.00 & 250\\
    
    \hline 

    \rowcolor{Gray}
    \textbf{Zero-shot} & & & & & & & & & & & &\\
    \hline

    Clip4Clip~\cite{Luo_Clip4Clip_2022} & \cmark & \xmark & 7.40 & 11.00 & 17.60 & 25.80 & 139.0 & 4.59 & 7.39 & 13.57 & 22.75 & 167.0\\
    Language-Bind~\cite{languagebind} & \cmark & \xmark & 2.60 & 4.60 & 9.00 & 17.20 & 190.5 & 2.20 & 4.20 & 8.20 & 15.80 & 204.5\\
    GestSync~\cite{Hegde23} & \xmark & \cmark & 3.60 & 5.60 & 13.20 & 19.80 & 212.5 & 3.20 & 6.60 & 18.40 & 29.80 & 127.5\\   
    \hline
    
    \rowcolor{Gray}
    \textbf{Fine-tuned} & & & & & & & & & & & &\\
    \hline

    Clip4Clip~\cite{Luo_Clip4Clip_2022} & \cmark & \xmark & 8.00 & 12.60 & 17.60 & 26.40 & 132.0 & 3.60 & 7.00 & 19.20 & 30.20 & 125.0\\
    Language-Bind~\cite{languagebind} & \cmark & \xmark & 5.80 & 10.80 & 14.00 & 20.40 & 140.5 & 4.80 & 8.00 & 12.60 & 24.40 & 180.0\\
    GestSync~\cite{Hegde23} & \xmark & \cmark & 10.00 & 18.20 & 27.40 & 41.20 & 70.5 & 11.60 & 16.60 & 27.40 & 40.00 & 82.5\\    GestureDiffuClip~\cite{ao2023gesturediffuclip} & \cmark & \xmark & 7.90 & 12.80 & 21.20 & 30.60 & 112.0 & 7.80 & 10.40 & 19.00 & 29.20 & 128.5\\ 
     
    \hline

    \rowcolor{Gray}
    \textbf{Ours} & & & & & & & & & & & &\\

    \hline

    \rowcolor{Blue}
    \modelname & \cmark & \xmark & 13.40 & 20.60 & 35.80 & 48.60 & 57.5 & 14.40 & 27.00 & 37.20 & 49.20 & 51.0\\
    \rowcolor{Blue}
    \modelname & \xmark & \cmark & 11.00 & 20.00 & 34.20 & 46.20 & 59.0 & 12.20 & 20.60 & 37.00 & 45.60 & 60.5\\
    \rowcolor{Blue}
    \modelname & \cmark & \cmark & \textbf{18.80} & \textbf{30.80} & \textbf{46.40} & \textbf{62.00} & \textbf{31.0} & \textbf{18.20} & \textbf{20.20} & \textbf{51.40} & \textbf{70.20} & \textbf{24.5}\\

  \bottomrule
  \end{tabular}
  \end{adjustbox}
  \vspace{-5pt}
\end{table*}

\subsection{AVSpeech Test Benchmarks}
\label{sec:test_datasets}
Using the AVSpeech official test set~\cite{Erphat_LookingToListen_2018}, we manually curate three separate evaluation benchmarks for the three downstream gesture tasks.
The statistics for the evaluation test sets are summarized in Table~\ref{tab:dataset}.

\vspace{3pt}
\noindent \textbf{AVS-Ret.} We create a new cross-modal retrieval benchmark containing diverse gesture clips of hundreds of unique speakers. We choose a gallery of $500$ clips, which also contain isolated clean speech and accurate text transcriptions. We verify that the clips contain reasonable gesture activity and transcripts with at least two nouns or verbs or adjectives. For evaluation, we use the standard metrics used in other video-text retrieval works~\cite{languagebind,wang2022internvideo}, i.e. Recall@K and Median Rank. We evaluate both gesture ($g$) to content (speech-text $c$) retrieval and vice-versa and show both unimodal and multimodal retrieval performance.

\vspace{3pt}
\noindent \textbf{AVS-Spot.} To quantitatively evaluate the gesture spotting task, we manually curate a new test dataset where we search and annotate clips that clearly contain a word that is gestured. We obtain $500$ such clips, each containing a target word that is clearly gestured. The manual annotation process removes all kinds of label noise in the test set, allowing for a faithful evaluation of our newly defined gesture spotting task. Additionally, we also manually annotate these target words with binary ``stress/emphasis" labels, which can have important cues about the gesture (Table~\ref{tab:stress}). We provide additional annotations of the AVS-Spot test set and more results with it in the supplementary.

\vspace{3pt}
\noindent \textbf{AVS-Asd.} To build the evaluation dataset for active speaker detection, we first choose $500$ ``target" clips. For these target clips, we create three evaluation subsets, where we choose $P - 1$ clips from different speakers, where $P = 2, 4, 6$. We report the accuracy of detecting the correct target speaker out of the $P$ different speakers.

\section{Results}
\label{sec:results}
\subsection{Baselines} For baselines, we report performance of zero-shot pre-trained vision-language models~\cite{Luo_Clip4Clip_2022, languagebind} and pre-trained GestSync~\cite{Hegde23}, which learns gesture-audio correspondences by solving for audio-visual synchronization. We also report scores after fine-tuning all these models further on our training data for a fair comparison. In addition, we compare with the semantic encoder of GestureDiffuCLIP~\cite{ao2023gesturediffuclip}, by training it on our dataset.

\subsection{Cross-modal Retrieval} 
In Table~\ref{tab:t2g}, we compare the performance of \modelname against other baselines on the cross-modal retrieval task. Zero-shot evaluation of foundational vision-language models like LanguageBind~\cite{languagebind} and Clip4Clip~\cite{Luo_Clip4Clip_2022} leads to higher than chance performance. These models are designed to capture different kinds of features: they cannot handle a large number of frames, and learn non-gesture attributes like identity and scene. Fine-tuning these models improves their performance on the task, but it is still far from the performance of~\modelname. GestSync~\cite{Hegde23} clearly performs better than the foundational vision-language models post-finetuning. However, since this network is trained to detect synchronization offsets in speech and video, its representations perform poorly for global semantic tasks like retrieval. This is also partly true for our model when we turn off the global phrase loss (Table~\ref{tab:loss} row $2$ vs row $3$). GestureDiffuCLIP's semantic encoder~\cite{ao2023gesturediffuclip} performs only second best among the baselines. The lack of local word-level semantic supervision leads to an inferior performance compared to~\modelname for both GestSync and GestureDiffuCLIP. 

Furthermore, none of the baseline approaches ingest and fuse multi-modal speech-text inputs. Our~\modelname model outperforms previous methods by a large margin.~\modelname can retrieve gestures from speech or text queries with similar performance. The opposite direction is also true, i.e.\ retrieving speech or text for a query gesture clip. Finally, retrieving with the fused speech-text representation is clearly better than the unimodal variants, showing that the speech and text embeddings each encode information that is not present in the other modality.

\begin{table}[h]
  \caption{Gesture-based word spotting performance on the AVS-Spot benchmark (Sec~\ref{sec:test_datasets}). }
  \label{tab:spotting}
  \centering
  \setlength{\tabcolsep}{11pt}
  \vspace{-5pt}
  \begin{tabular}{l|c|c|c}
    \toprule

    \multirow{2}{*}{\textbf{Method}} & \multicolumn{2}{c|}{\textbf{Mod.}} & \multirow{2}{*}{\textbf{Accuracy} $\uparrow$}\\\cline{2-3}
    
    & \textbf{T} & \textbf{A}  \\
    
    \midrule

    \hline
    \rowcolor{Gray}
    \textbf{Zero-shot} & &  & \\
    \hline

    Clip4Clip~\cite{Luo_Clip4Clip_2022} & \cmark & \xmark & 9.20 \\
    Language-Bind~\cite{languagebind} & \cmark & \xmark & 9.40 \\
    GestSync~\cite{Hegde23} & \xmark & \cmark & 15.63 \\

    \hline
    
    \rowcolor{Gray}
    \textbf{Fine-tuned} & & &  \\
    \hline

    Clip4Clip~\cite{Luo_Clip4Clip_2022} & \cmark & \xmark & 17.59 \\
    Language-Bind~\cite{languagebind} & \cmark & \xmark & 15.80\\
    GestSync~\cite{Hegde23} & \xmark & \cmark & 21.04 \\    GestureDiffuClip~\cite{ao2023gesturediffuclip} & \cmark & \xmark & 19.50 \\
    
    \hline

    \rowcolor{Gray}
    \textbf{Ours} & & &  \\

    \hline

    \rowcolor{Blue}
    \modelname & \cmark & \xmark & 61.00 \\
    \rowcolor{Blue}
    \modelname & \xmark & \cmark & 41.80 \\
    \rowcolor{Blue}
    \modelname & \cmark & \cmark & \textbf{63.60} \\

  \bottomrule
  \end{tabular}
  \vspace{-5pt}
\end{table}

\subsection{Gesture Word Spotting} 
Table~\ref{tab:spotting} compares the spotting accuracy of different methods on the AVS-Spot benchmark. Unlike~\modelname that uses the word-level $L_{couple}$ loss, all the baseline methods, including the semantic encoder of GestureDiffuCLIP~\cite{ao2023gesturediffuclip}, use only a phrase-level loss and hence, struggle to learn fine-grained word-level associations. Through this task, we also see the big advantage of using text modality in addition to speech -- text-based gesture spotting is more accurate than using audio. This is expected, since word-level semantic correspondences are easier to learn in text space. Having said that, using speech alongside text still gives a clear improvement even in the gesture spotting task.

\begin{figure*}[h]
\includegraphics[width=\linewidth]{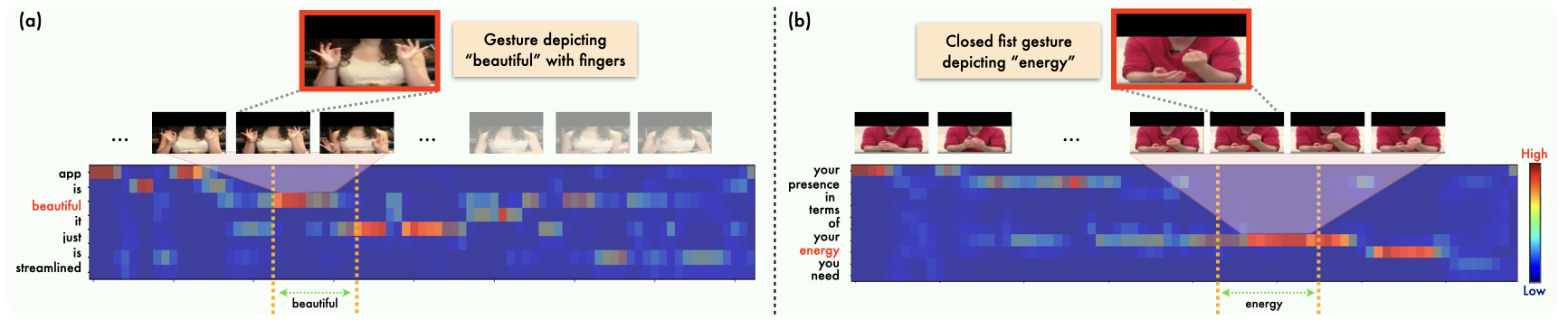}
  \vspace{-20pt}
  \caption{JEGAL can spot the gestured words in a video clip. Here, we show a similarity heatmap of words vs video frames. The vertical yellow lines indicate the speech-based word boundaries of `beautiful' and `energy'. The red triangles zoom into the corresponding frames where JEGAL detects the words, clearly aligning with the gestures. For the word ``beautiful'' the gestured segment is smaller than the spoken word boundary and for the word ``energy'' it extends well beyond to the right of the word boundary. The alignment layer (Sec~\ref{subsec:alignment}) allows the model to look beyond just the speech-based boundaries. Note that our model learns to perform gestured-word spotting without using any training labels on which words are gestured.}
  \label{fig:spotting}
  \vspace{-14pt}
\end{figure*}

In Fig~\ref{fig:spotting}, we show two examples of spotting gestured words. The heatmaps show the similarity of the word (red is higher) along the video frames. In the first example, the lady gestures ``beautiful'' with her fingers, and in the second example, the speaker clenches the fist to show ``energy''. We can see that not all words get a high similarity score, only ones with distinctive gestures are spotted. Furthermore, the spotting does not exactly align with the speech-based word boundary (indicated by yellow vertical lines). Our alignment layer (Sec~\ref{subsec:alignment}) allows the model to look beyond the speech-based boundary and find the exact frames where the word is gestured.

\subsection{Active Speaker Detection}
In Table~\ref{tab:asd}, we show the accuracy of identifying the target speaker for a given text and/or speech segment. This task is different from our other tasks -- it does not need a strong holistic understanding of the gesture sequence like retrieval, nor does it need semantic word-level understanding. It can simply be solved by checking for frame-level synchronization, which is exactly why we see GestSync~\cite{Hegde23} performs the best on this task. None of the other models are trained with strong frame-level video-speech supervision, and hence, perform worse.~\modelname comes at a close second after GestSync. We also see that speech information is more useful here compared to text.

\begin{table}[h]
  \caption{Performance of active speaker detection on AVS-Asd benchmark. We report the mean class accuracy of predicting the active speaker among a set of $S$ speakers, where $S=2, 4, 6$. }
  \label{tab:asd}
  \centering
  \setlength{\tabcolsep}{5pt}
  \vspace{-5pt}
  \begin{tabular}{l|c|c|c|c|c}
    \toprule
    \multirow{2}{*}{\textbf{Method}} & \multicolumn{2}{c|}{\textbf{Mod.}} & \multirow{2}{*}{\textbf{2 spk.}} & \multirow{2}{*}{\textbf{4 spk.}} & \multirow{2}{*}{\textbf{6 spk.}} \\\cline{2-3}
    & \textbf{T} & \textbf{A} &\\
    \midrule
    Random & \cmark & \cmark & 50.0 & 25.0 & 16.7  \\

    \hline
    \rowcolor{Gray} 
    \textbf{Zero-shot} & & & & &\\
    \hline
    
    Clip4Clip~\cite{Luo_Clip4Clip_2022} & \cmark & \xmark & 62.6 & 39.6 & 31.8\\
    Language-Bind~\cite{languagebind} & \cmark & \xmark & 51.4 & 32.2 & 23.8\\
    GestSync~\cite{Hegde23} & \xmark & \cmark & 54.2 & 31.6 & 23.8\\

    \hline
    \rowcolor{Gray} 
    \textbf{Fine-tuned} & & & & &\\
    \hline

    Clip4Clip~\cite{Luo_Clip4Clip_2022} & \cmark & \xmark & 63.8 & 42.3 & 32.8\\
    Language-Bind~\cite{languagebind} & \cmark & \xmark & 56.8 & 38.7 & 30.1 \\
    GestSync~\cite{Hegde23} & \xmark & \cmark & \textbf{81.2} & \textbf{64.8} & \textbf{54.4}\\    GestureDiffuClip~\cite{ao2023gesturediffuclip} & \cmark & \xmark & 61.8 & 39.4 & 28.6\\

    \hline
    \rowcolor{Gray}
    \textbf{Ours} & & & & & \\

    \hline

    \rowcolor{Blue}
    \modelname & \cmark & \xmark & 65.6 & 44.4 & 34.6\\
    \rowcolor{Blue}
    \modelname & \xmark & \cmark & 74.4 & 50.0 & 40.4\\
    \rowcolor{Blue}
    \modelname & \cmark & \cmark & \textbf{76.8} & \textbf{57.8} & \textbf{48.0}\\
    
  \bottomrule
  \end{tabular}
  \vspace{-10pt}
\end{table}

\section{Insights and Ablations}
In this section, we provide additional insights into the gesture signals learned from the speech and text modalities, the impact of the two loss functions on the downstream tasks, and the choice of speech-text fusion.

\subsection{Speech v/s Text Modalities}
\label{subsec:speechVStext}
\begin{figure}[h]
\centering
\includegraphics[width=\linewidth]{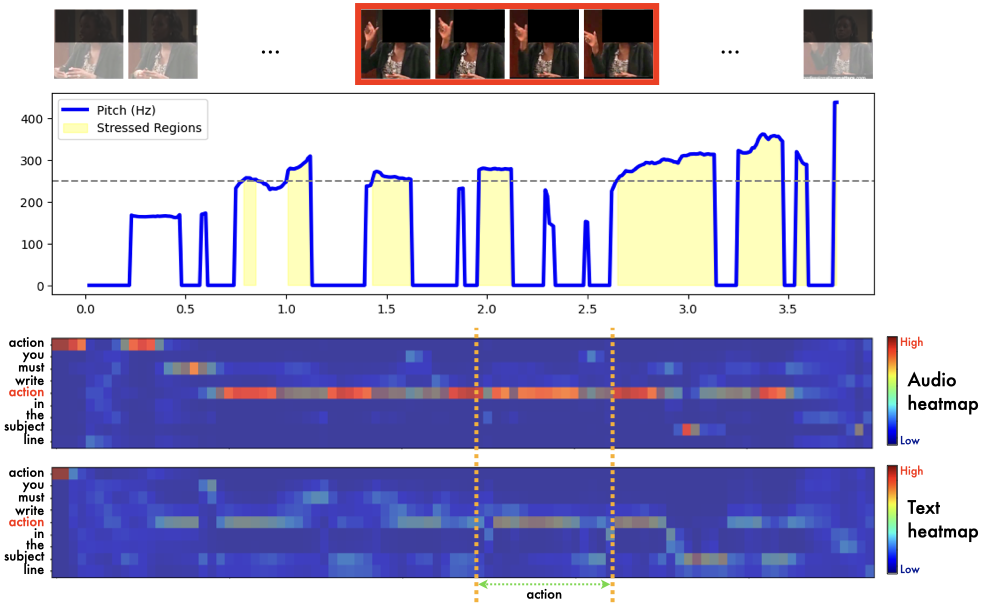}
  \vspace{-20pt}
  \caption{Audio and text heatmap examples when words are stressed.}
  \label{fig:stress}
  \vspace{-10pt}
\end{figure}

We have already seen how our model can flexibly leverage speech or text modality to solve three different kinds of tasks. In Fig~\ref{fig:stress}, we show an example where audio-based gesture word spotting is successful, but text-based spotting is not. We see that the uttered word ``action" has been emphasized (using pitch graph). This example leads us to do a deeper analysis of gesture word spotting based on stress cues. We divide the binary stress labels to split the AVS-Spot test set ($500$ samples) into two subsets, one containing only samples with stressed/emphasized words ($100$ samples) and the other subset containing the remaining words ($400$ samples). In Table~\ref{tab:stress}, we report the same spotting accuracy metric on these subsets separately. While we saw previously in Table~\ref{tab:spotting} that speech-based gesture spotting is clearly inferior compared to text-based gesture spotting, it is not always the case. We see that the difference in spotting accuracy between the stressed and non-stressed words is higher for the speech modality. This indicates that speech modality pays more attention to emphasis of the word compared to the text modality. 

\begin{table}[h!]
  \caption{Stressed words are more likely to be spotted with speech-based spotting than non-stressed words. As seen in the column ``$\Delta$'', the difference between stressed vs. non-stressed word spotting is higher for the speech modality.}
  \label{tab:stress}
  \centering
  \setlength{\tabcolsep}{5pt}
  \vspace{-8pt}
  \begin{tabular}{l|c|c|c|c}
    \toprule
    \textbf{Modality} & \textbf{With stress} & \textbf{W/o stress} & \textbf{All words} & $\Delta$\\
    \midrule
    Text & 68.00 & 59.25 & 61.00 & 14.8\% \\
    Speech & 54.00 & 38.75 & 41.80 & \cellcolor{green!12}39.4\% \\
  \bottomrule
  \end{tabular}
  \vspace{-8pt}
\end{table}

\subsection{Impact of loss functions}

In Table~\ref{tab:loss}, we study the impact of our two loss functions. In the first row, we only train with the phrase contrastive loss, which captures the global semantics. The second row is the variant trained only with gesture-word coupling loss, which captures local word-level semantics. Compared to the dual loss model, the results from these individual loss models are significantly worse. Specifically, the variant without the word coupling loss performs poorly on the spotting task and the one without the sequence contrastive loss performs poorly on retrieval and active speaker detection. The combination of the two losses performs the best across all the tasks, thus demonstrating the complementary nature of the two training objectives.

\begin{table}[h!]
  \caption{Each loss encourages feature learning at different temporal granularity, and combination of the two loss functions performs the best. 
  }
  \label{tab:loss}
  \centering
  \setlength{\tabcolsep}{1.8pt}
  \vspace{-10pt}
\begin{adjustbox}{width=1.0\columnwidth,center}
  \begin{tabular}{l|c|c|c|c|c}
    \toprule
    \multirow{2}{*}{\textbf{Loss}} & \multicolumn{3}{c|}{\textbf{Retrieval}} & \textbf{Spotting} & \textbf{ASD}\\\cline{2-6}
         & \textbf{R@5} $\uparrow$ & \textbf{R@10} $\uparrow$ & \textbf{MR} $\downarrow$ & \textbf{Acc.} $\uparrow$ & \textbf{Acc.} $\uparrow$\\
    \midrule
    Seq. contrastive  & 12.20 & 23.60 & 45 & 20.83 & 44.2\\
    Word coupling  & 8.50 & 14.60 & 76 & 52.46 & 14.8\\
    \rowcolor{Blue}
    Seq. + Word coupling & \textbf{18.80} & \textbf{30.80} & \textbf{31} & \textbf{63.60} & \textbf{48.0}\\
    
  \bottomrule
  \end{tabular}
  \end{adjustbox}
  \vspace{-12pt}
\end{table}

\subsection{Fusion techniques}
In Table~\ref{tab:fusion}, we ablate different ways to fuse the speech and text features. The first case is to not fuse at all and have two separate pairwise contrastive losses: gesture-audio and gesture-text. We can see in the first two rows that this is an inferior design. In fact, it is better to train with a single contrastive head after fusing the speech and text embeddings (rows $3$, $4$). The fusion strategy of choice would be to concatenate, rather than average.

\vspace{-5pt}
\begin{table}[h!]
  \caption{ 
Ablation study on fusing speech and text modalities. Training without fusing is far worse, as the model cannot perform the tasks by using multiple information streams at the same time.
  }
  \label{tab:fusion}
  \centering
  \setlength{\tabcolsep}{1.8pt}
  \vspace{-10pt}
\begin{adjustbox}{width=1.0\columnwidth,center}
  \begin{tabular}{l|c|c|c|c|c}
    \toprule
    & \multicolumn{3}{c|}{\textbf{Retrieval}} & \textbf{Spotting} & \textbf{ASD}\\\cline{2-6}
    & \textbf{R@5} $\uparrow$ & \textbf{R@10} $\uparrow$ & \textbf{MR} $\downarrow$ & \textbf{Acc.} $\uparrow$ & \textbf{Acc.} $\uparrow$\\
    
    \midrule

    Pairwise (with text) & 9.39 & 15.58 & 70 & 34.31 & 29.6\\
    Pairwise (with audio) & 9.80 & 16.60 & 72 & 23.67 & 31.4\\
    \hline
    Late fusion (avg.) & 17.00 & 26.40 & 40 & 56.04 & 41.2\\
    \rowcolor{Blue}
    Late fusion (concat.) & \textbf{18.80} & \textbf{30.80} & \textbf{31} & \textbf{63.60} & \textbf{48.0}\\
    
  \bottomrule
  \end{tabular}
  \end{adjustbox}
  \vspace{-15pt}
\end{table}

\section{Conclusion}
In this work, we learn a joint embedding space that captures cross-modal relationships with gestures, speech, and language. We show that we can learn such an embedding space with weak supervision using a careful design of two loss functions. We evaluate these new representations on three new downstream tasks and manually curated test sets. We observe that the two modalities, i.e., speech and text, learn complementary features that can be useful for different kinds of gesture-related tasks. One promising future direction would be to explore 2D and 3D keypoint-based inputs to make the network computationally lighter and less susceptible to distracting features. 

\vspace{2mm}
\noindent
\textbf{Acknowledgements.}
The authors would like to thank Piyush Bagad, Ragav Sachdeva, Jaesung Hugh, Paul Engstler for their valuable discussions. The authors are further grateful to Alyosha Efros, Jitendra Malik, and Justine Cassell for their insightful inputs and suggestions. They also extend their thanks to David Pinto for setting up the data annotation tool and to Ashish Thandavan for his support with the infrastructure. This research is funded by EPSRC Programme Grant VisualAI EP/T028572/1, an SNSF Postdoc.Mobility Fellowship P500PT\_225450 and a Royal Society Research Professorship RSRP\textbackslash R\textbackslash 241003.

{
    \small
    \bibliographystyle{ieeenat_fullname}
    \bibliography{main, shortstrings, vgg_local}

}

\newpage
\clearpage
\appendix

\section{Most Gestured Words}
\begin{figure}[h]
\includegraphics[width=\linewidth]{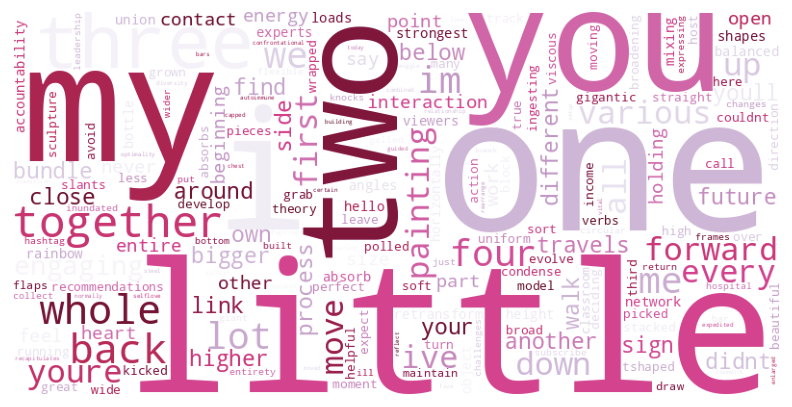}
\vspace{-20pt}
  \caption{
Word cloud for the most commonly gestured words.
  }
  \label{fig:wordcloud}
  \vspace{-15pt}
\end{figure}

In Figure~\ref{fig:wordcloud}, we show the most commonly spotted gestured words that are spotted by~\modelname on the AVS-Spot test set: pointing gestures (you, my, we), adjectives/adverbs (little, open, whole, gigantic, broad), direction words (forward, here, below) and numbers (one, two, first). 

\section{Additional Evaluations and Analysis}

\subsection{Gesture Word Spotting: Evaluation in challenging conditions}

\begin{figure*}[!t]
    \centering{\includegraphics[width=\linewidth]{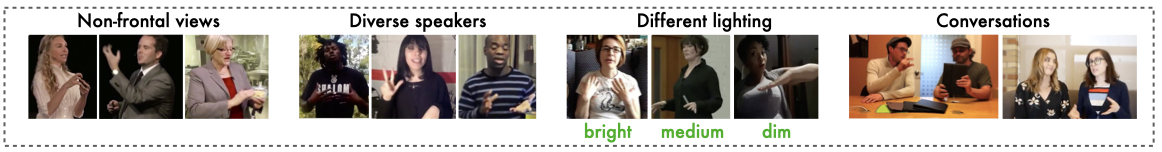}}
    \caption{The AVS-Spot test set is quite diverse -- some examples are shown above. Additionally, we annotate the clips in AVS-Spot for frontal/non-frontal views and lighting and analyze the performance on these individual subsets.}
    \label{fig:data_diversity}
\end{figure*}

Our evaluation set (constructed from AVSpeech) includes a diverse range of samples: (i) non-frontal videos, (ii) varying lighting conditions, (iii) a wide variety of speakers, and (iv) conversational videos (from which we extract segments featuring a single speaker). In this section, we specifically benchmark the performance of JEGAL on these challenging subsets. We label the AVS-Spot dataset with new metadata: (i) lighting conditions (dim, medium, bright), and (ii) speaker poses (frontal vs. non-frontal). Fig~\ref{fig:data_diversity} illustrates the diversity of the test set. 

Table~\ref{table:challenging_conditions} reports the spotting accuracy across these subsets. We find that the model performs best on brightly lit videos, with similar accuracy for dim and medium lighting. 

\begin{table}[!th]
\centering
    \small
    \setlength{\tabcolsep}{3pt}
    \vspace{-5pt}
    \caption{Evaluation in challenging conditions: JEGAL outperforms prior models in all settings.}
    \vspace{-10pt}
    \begin{tabular}{c||c|c|c||c|c}
    \hline

    \multirow{2}{*}{\textbf{Method}} & \multicolumn{3}{c||}{\textbf{Lighting}} & \multicolumn{2}{c}{\textbf{Speaker pose}} \\\cline{2-6}

    & \textbf{Dim} & \textbf{Medium} & \textbf{Bright} & \textbf{Frontal} & \textbf{Non-frontal} \\
    
    \hline

    GestSync~\cite{Hegde23} & 9.67 & 22.92 & 8.33 & 21.59 & 17.80\\
    GestDiffuClip~\cite{ao2023gesturediffuclip} & 15.6 & 19.7 & 21.8 & 19.35 & 19.90\\
    
    \hline
    
    \rowcolor{Blue}
    \textbf{JEGAL (Ours)} & 61.29 & 62.58 & 77.77 & 62.76 & 68.49\\
    
    \hline
    \end{tabular}
    \label{table:challenging_conditions}
\end{table}

\subsection{Effect of Modality dropping} 
We present the impact of dropping text and audio modalities at varying rates on the spotting task in Table~\ref{table:modality_drop}. A drop rate of $30\%$ means that during training, either text or audio is randomly dropped in $30\%$ of the batch samples. Dropping the modalities at $50\%$ performs the best across all inference-time settings.

\begin{table}[ht]
    \centering
    \small
    \setlength{\tabcolsep}{16pt}
    \caption{Dropping modalities evenly during training works best.}
    \vspace{-10pt}
    \begin{tabular}{c|c|c|c}
    \hline

    \multirow{2}{*}{\textbf{Drop \%}} & \multicolumn{3}{c}{\textbf{Accuracy} $\uparrow$}\\\cline{2-4}

    & \textbf{T} & \textbf{A} & \textbf{TA} \\

    \hline
    
     30\% & 52.2 & 38.6 & 63.2\\ 
     \rowcolor{Blue}
     50\% (JEGAL) & 61.0 & 41.8 & 63.6\\
     70\% & 61.3 & 42.2 & 62.6\\

    \hline
    \end{tabular}
    \vspace{-10pt}
    \label{table:modality_drop}
\end{table} 

\subsection{Computational efficiency} Table~\ref{table:computation} shows the inference time (averaged across ten runs) for a $5$-second input on a single NVIDIA $V100$ GPU. Our model can process $\approx 52$ frames per second, indicating that the inference is quite fast but is not streaming-capable yet, as the bidirectional transformer attends to all future frames provided as context.

\vspace{-8pt}
\begin{table}[ht]
    \centering
    \small
    \setlength{\tabcolsep}{4pt}
    \caption{Inference time analysis for 5-second input.}
    \vspace{-10pt}
    \begin{tabular}{c||c|c|c|c}
    \hline

     & \textbf{Visual Enc.} & \textbf{Text Enc.} & \textbf{Audio Enc.} & \textbf{Total} \\
    \hline

    inf. time (sec) $\downarrow$ & 1.84 & 0.18 & 0.07 & 2.09 \\
    
    \hline
    \end{tabular}
    \vspace{-7pt}
    \label{table:computation}
\end{table}

\subsection{Where does the model focus on?}

\begin{figure*}[!t]
    \centering{\includegraphics[width=\linewidth]{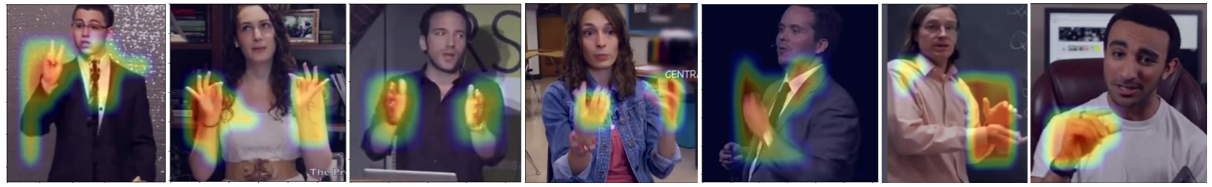}}
    \caption{We plot the activation maps of the visual features of JEGAL. We can see that JEGAL focuses strongly on the hand gestures.}
    \label{fig:act_maps}
\end{figure*}

We visualize the activation maps of the visual features of JEGAL to see which spatial region of the video the model focuses on. In Fig~\ref{fig:act_maps}, we see that the model focuses on the hand gestures. 

\section{Model Details}
In Table~\ref{tab:arch}, we provide detailed description of the model architecture. The code and models have been released to support future research.

\begin{table*}[h]
\centering
\caption{Overview of the model architecture, detailing the input modalities, network components, and key parameters used in each stage of our framework.}
\label{tab:arch}
\setlength{\tabcolsep}{8pt}
\begin{tabular}{|l|l|l|l|}
\hline
\textbf{Branch} & \textbf{Layer/Module} & \textbf{Input Shape} & \textbf{Output Shape} \\
\hline
\multicolumn{4}{|l|}{\textbf{Visual Branch}} \\
\hline
& Vision backbone & 3 × T × 270 x 480 & T × 1024 \\
\cline{2-4}
& Projection MLP & & \\
& - Linear & T × 1024 & T × 512 \\
& - LayerNorm & T × 512 & T × 512 \\
& - ReLU & T × 512 & T × 512 \\
& - Linear & T × 512 & T × 512 \\
\cline{2-4}
& Positional Encoding & T × 512 & T × 512 \\
\cline{2-4}
& Transformer (N=6 layers) & & \\
& - Self-Attention (h=8) & T × 512 & T × 512 \\
& - Feed Forward & T × 512 & T × 512 \\
\cline{2-4}
& Output Projection & T × 512 & T × 512 \\
\hline
\multicolumn{4}{|l|}{\textbf{Text Branch}} \\
\hline
& mRoberta Text backbone & W & W × 768 \\
\cline{2-4}
& Transformer (N=3 layers) & & \\
& - Self-Attention (h=8) & W × 768 & W × 768 \\
& - Feed Forward & W × 768 & W × 768 \\
\cline{2-4}
& Output Projection & W × 768 & W × 256 \\
\hline
\multicolumn{4}{|l|}{\textbf{Audio Branch}} \\
\hline
& Melspectrogram Input & 1 × 80 × 4T & - \\
\cline{2-4}
& Conv2D + BN + ReLU & & \\
& (k=5, s=1, p=2) & 1 × 80 × 4T & 32 × 80 × 4T \\
\cline{2-4}
& Conv2D + BN + ReLU & & \\
& (k=3, s=2, p=1) & 32 × 40 × 2T & 64 × 40 × 2T \\
\cline{2-4}
& Conv2D + BN + ReLU & & \\
& (k=3, s=2, p=1) & 64 × 40 × 2T & 128 × 20 × T \\
\cline{2-4}
& Conv2D + BN + ReLU & & \\
& (k=3, s=(3,1), p=1) & 128 × 7 × T & 256 × 7 × T \\
\cline{2-4}
& Conv2D + BN + ReLU & & \\
& (k=3, s=(3,1), p=1) & 256 × 3 × T & 256 × 3 × T \\
\cline{2-4}
& Conv2D & & \\
& (k=1, s=(3,1), p=0) & 256 × 3 × T & 256 × 1 × T \\
\cline{2-4}
& Output Projection + reshape & 256 × 1 × T & T × 256 \\
\hline
\multicolumn{4}{|l|}{\textbf{Late Fusion}} \\
\hline
& Encoded Features & & \\
& - Visual & T × 512 & - \\
& - Text + sub-word pooling & W × 256 & W × 256 \\
& - Audio + sub-word pooling & T × 256 & W × 256 \\
\hline
\end{tabular}
\end{table*}

\section{Dataset Visualization}
\begin{figure*}[ht]
\center{
\includegraphics[width=\linewidth]{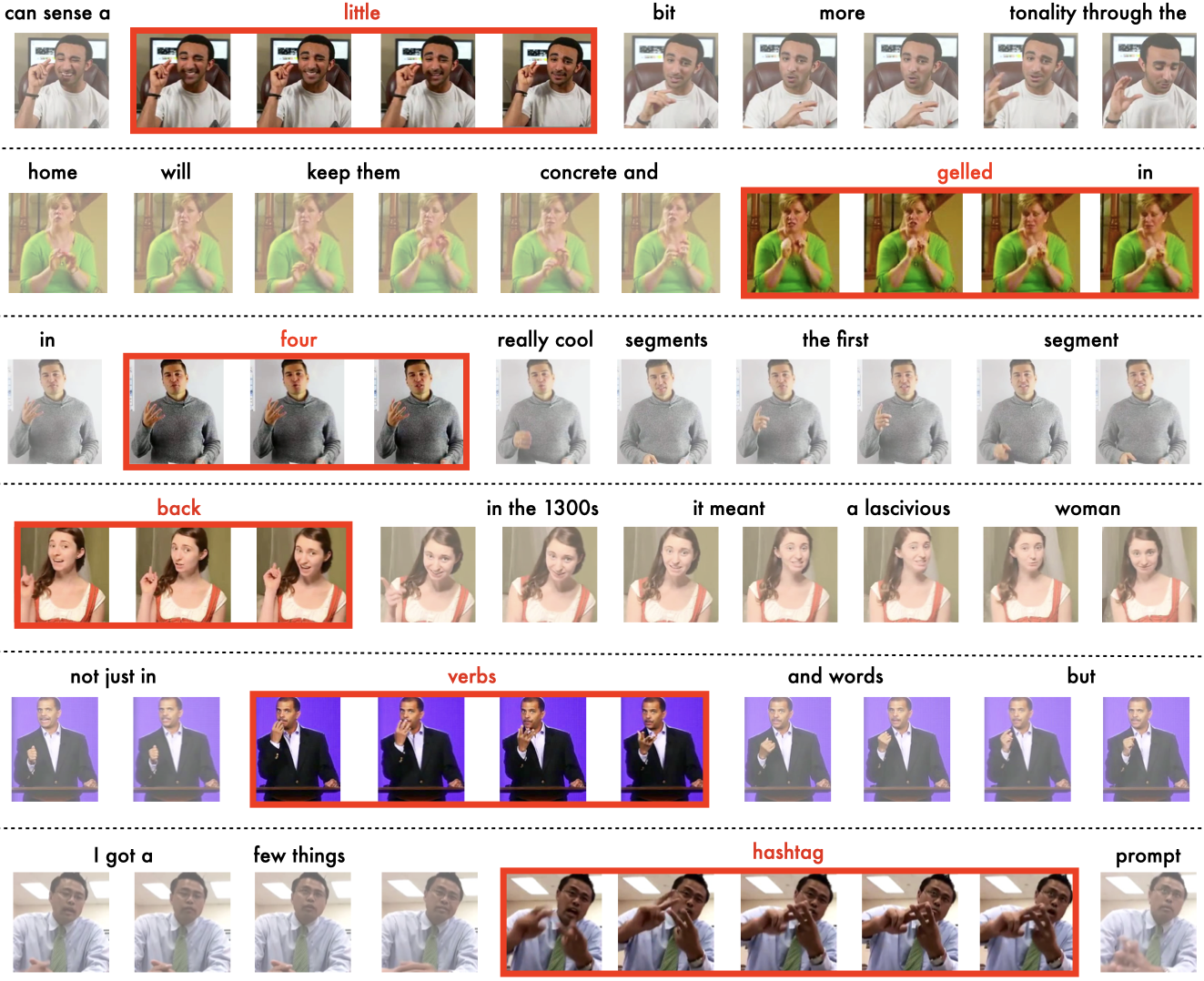}
  \caption{Visualization of the \textbf{AVS-Spot} dataset, showcasing video frames from different samples. Each row corresponds to a single video, with the highlighted keyword indicating the annotated gestured word for spotting. The figure illustrates the dataset's diversity, featuring a wide range of unique keywords, various speakers, and distinct gestures.}
  \label{fig:spotting_samples}
  }
\end{figure*}

In Figure~\ref{fig:spotting_samples}, we present examples from our manually annotated AVS-Spot test set (curated from the publicly available AVSpeech test dataset~\cite{Erphat_LookingToListen_2018}), designed to evaluate downstream gesture spotting performance. As shown, the dataset includes a diverse collection of unique words, carefully curated to ensure clear and contextually appropriate gestures. For instance, in row-$1$, the word ``little'' is accompanied by a gesture where two fingers move close together to indicate a small size; in row-$4$, the speaker points backward to represent the word ``back''; and in row-$6$, the fingers of both hands move in a distinctive pattern to indicate ``hashtag''.

\section{Qualitative Results}
\begin{figure*}[ht]
\center{
\includegraphics[width=7in, height=5.3in]{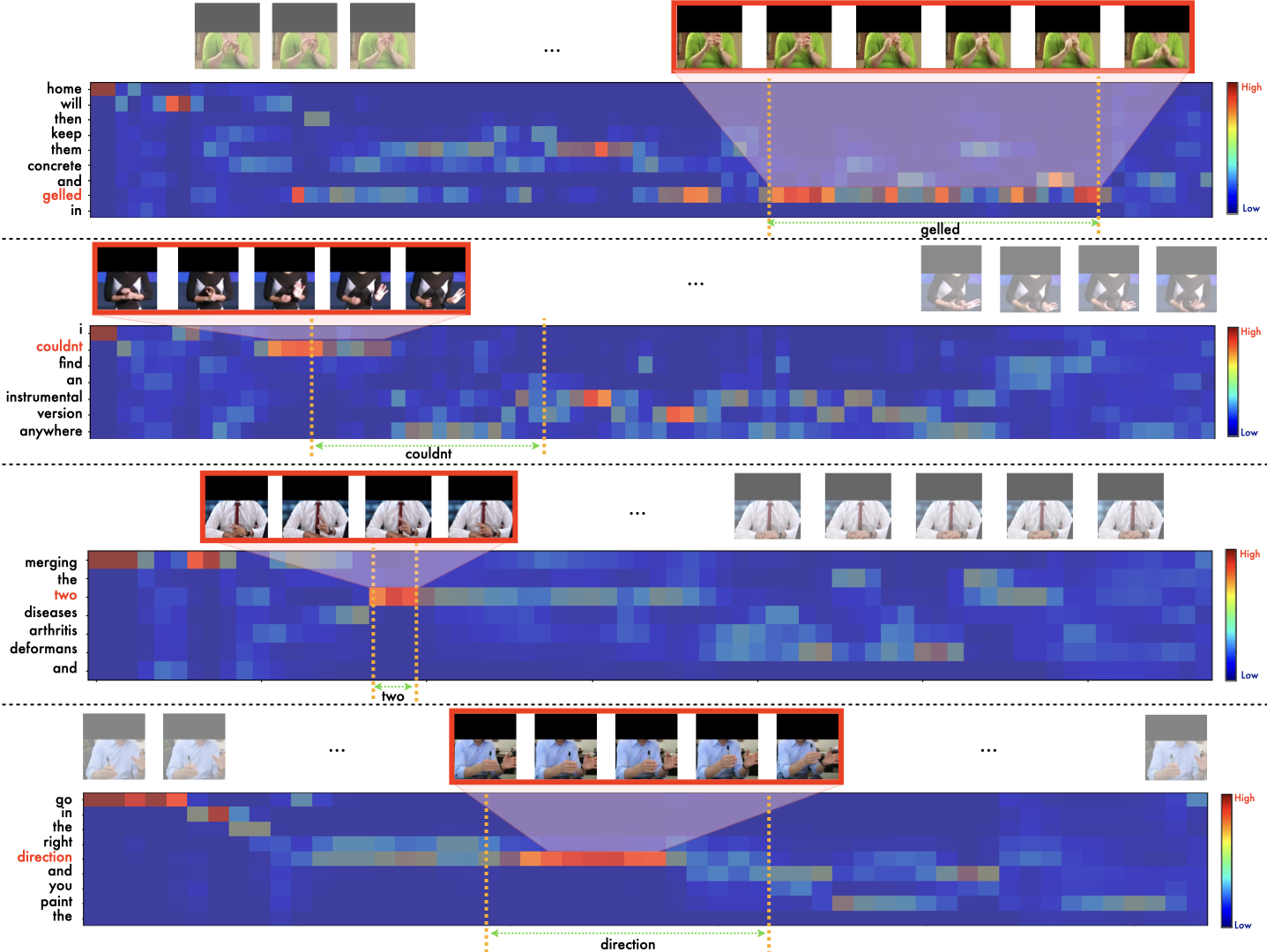}
  \caption{Additional gestured word spotting results on AVS-Spot dataset. Keywords are highlighted in red on the left panel and the speech-based force alignment word boundaries are marked by yellow lines. JEGAL successfully spots the gestured keywords, demonstrating its robustness across diverse gestures and speakers. The red triangles zoom into the corresponding frames where JEGAL detects the keywords, clearly aligning with the gestures. Note that in some cases (e.g., rows $2$ and $4$), ground-truth boundaries may slightly differ, as the speaker can gesture and utter the same word at slightly different times. JEGAL effectively estimates the approximate intervals where the target word is gestured.}
  \label{fig:additional_spotting}
  }
\end{figure*}

In Fig~\ref{fig:additional_spotting}, we show additional qualitative examples for gesture spotting. In the left text panel, the red-highlighted word represents the keyword to be spotted, as curated in the AVS-Spot test set. The word-labeled vertical columns, separated by yellow lines, indicate the word boundaries derived from speech-text alignment. JEGAL successfully spots most of these keywords, as shown by the red heatmaps. Notably, the boundaries may vary slightly since speakers often gesture and speak at slightly different times, highlighting the inherent challenges of our weakly-supervised gesture representation learning task.

\begin{figure*}[ht]
\center{
\includegraphics[width=\linewidth]{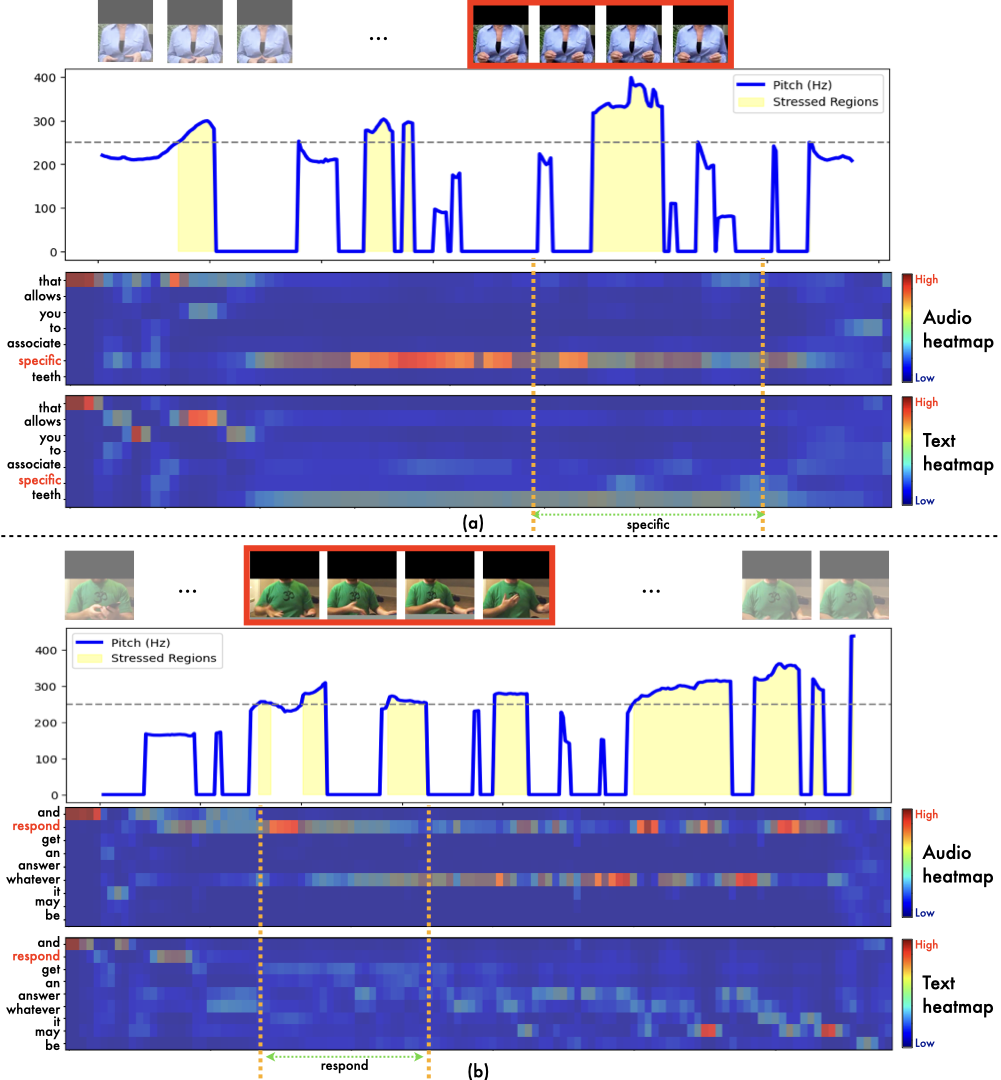}
  \caption{Examples highlighting the role of stressed speech regions in audio-based gesture spotting. The audio-only model successfully detects the stressed keywords ``specific'' and ``respond'', whereas the text-only model misses these words. Evidently, the audio-only model looks for word emphasis cues (indicated by high pitch) as such words are more likely to be gestured. This would be difficult to infer from text modality alone. These examples illustrate the advantages of leveraging audio cues for gesture spotting.}
  \label{fig:additional_stress}
  }
\end{figure*}

In Fig~\ref{fig:additional_stress}, we present additional examples demonstrating that audio-based gesture spotting tends to focus on ``stressed regions'' in speech, unlike text-based spotting. This difference is evident from the audio and text heatmaps for each sample. In Fig~\ref{fig:additional_stress}, our model detects the stressed keywords ``specific'' and ``respond'', whereas the text-only model misses these words. Evidently, the audio-only model looks for word emphasis cues (indicated by high pitch) as such words are more likely to be gestured. This would be difficult to infer from text modality alone. These examples illustrate the advantages of leveraging audio cues for gesture spotting.

\section{Limitations and Areas of Improvement}
Our work is the first to tackle large-scale co-speech gesture understanding. We highlight some of the limitations of our approach here. One aspect the model struggles with is when there are limited gesture actions or hand movements that are unrelated to speech. Another shortcoming is that since we learn with only weak sequence-level supervision, the model can ``find shortcuts" by focusing on simple rhythmic hand movements that occur in certain gestures classes like the beat gestures. This can affect the representation quality of iconic and deitic gestures that contain clear semantic meaning. While we still show that our models can spot such gestures, future works can focus on improving this imbalance in gesture classes.

\section{Potential Negative Societal Impacts}
While our research significantly contributes to advancing gesture understanding, there are some potential risks of surveillance, as the system could infer conversations from a distance by identifying words/phrases. Nonetheless, we believe the benefits outweigh these risks, as the technology enhances human-machine interaction by integrating non-verbal cues. According to the $55$\% rule~\cite{55rule}, non-verbal communication constitutes $55$\% of overall communication. This highlights the importance of enabling machines to engage in holistic, natural interactions with humans by understanding non-verbal elements like gestures.

\end{document}

%% file: preamble.tex
\usepackage[subtle]{savetrees} 